\documentclass[dvips]{llncs}
\usepackage{amsmath, amsfonts, amssymb, dsfont}
\usepackage{graphicx}

\title{Evolutionary multi-stage financial scenario tree generation}
\author{Ronald Hochreiter}
\institute{Department of Finance, Accounting and Statistics\\WU Vienna University of Economics and Business\\ \email{ronald.hochreiter@wu.ac.at}}

\begin{document}

\maketitle

\begin{abstract}
Multi-stage financial decision optimization under uncertainty depends on a careful numerical approximation of the underlying stochastic process, which describes the future returns of the selected assets or asset categories. Various approaches towards an optimal generation of discrete-time, discrete-state approximations (represented as scenario trees) have been suggested in the literature. In this paper, a new evolutionary algorithm to create scenario trees for multi-stage financial optimization models will be presented. Numerical results and implementation details conclude the paper.\\ \par
{\bf Key words:} Optimization under uncertainty, scenario generation, scenario optimization, financial decision theory, risk management
\end{abstract}

\section{Introduction}
\label{sec:introduction}

Stochastic programming is a versatile method to model and solve decision problems under uncertainty. See \cite{RuszczynskiShapiro2003} for an overview of the area of stochastic programming, and \cite{WallaceZiemba2005} for stochastic programming languages, environments, and applications. 

We consider the following generalized formulation of a multi-stage stochastic financial optimization model. The decision taker faces a discrete-time decision horizon $t = 1, \ldots, T$, and a set of investment assets (or asset categories) $\mathcal{A}$ with uncertain future returns $V_a$. These uncertain returns are represented by a stochastic process discretized into a multi-variate, multi-stage scenario tree. This scenario tree is used to build either a deterministic equivalent model formulation, which can be solved using off-the-shelf solvers, or to use a stochastic decomposition algorithm to obtain numerical solutions of the problem. 
The objective function consists of a risk-return bi-criteria functional, whereby the aim is to maximize the expected wealth and to minimize some risk functional $\mathds{F}$ of the wealth at the terminal stage $T$. This resembles the classical Markowitz-style asset allocation \cite{Markowitz1952}, see also the multi-stage generalization presented by \cite{Steinbach2001}. The chosen risk factor does not necessarily have to be the variance, e.g. other coherent risk measures as shown by \cite{ArtznerEtAl1999} or similar probability-based measures might be better suited for different risk management purposes, and can be integrated into the model. Both dimensions - expectation and risk - are weighted using a risk-aversion parameter $\kappa$, which can be adapted to the needs of the investor and to the current market situation. The main decision is concerned with the amount of budget $b_a$ to be invested into each asset (or asset category) $a$, as the portfolio is rebalanced at each stage $t = 2, \ldots, T-1$. There is no rebalancing at terminal stage $T$. Furthermore, additional investment budget $B$ is available at each stage up to $T-1$, which is deterministically determined in advance in this basic model. An important constraint is that the amount of purchases $p$ in each stage cannot exceed the sum of the amount of sales $s$ plus the additional budget available at the respective stage.

Given the above problem specification, we may formulate our multi-stage stochastic programming model as shown in Eq. \ref{equ:msam}. The numbers in square brackets represent the stage(s) at which the respective constraint is active. \begin{eqnarray}
\begin{array}{llll}
\mbox{maximize} & \mathds{E} ( \sum_{a \in \mathcal{A}} b_a, T ) +  \kappa \mathds{F} ( \sum_{a \in \mathcal{A}} b_a, T ) & & \\
\mbox{subject to} & \sum_{a \in \mathcal{A}} b_{a} = B & & [1] \\
&  b_{a} \leq V_a b_{a}^{(-1)} + p_{a} - s_{a} & \forall a \in \mathcal{A} & [2, \ldots, T-1] \\
& \sum_{a \in \mathcal{A}} p_{a} \leq \sum_a s_{a} + B & & [2, \ldots, T-1] \\
&  b_{a} \leq V_a b_{a}^{(-1)}  & \forall \mathcal{A}  & [T] \\
& b_a, p_a, s_a \geq 0 & \forall \mathcal{A} & [1, \ldots, T] \\
\end{array}
\label{equ:msam}
\end{eqnarray}
The multi-stage recourse decision can be observed in the second and the fourth constraint: $V_a$ represents the future asset return of asset (or asset category) $a$ in the respective stage (on the scenario tree) and is multiplied by the invested budget $b_{a}^{(-1)}$ of the previous stage.

The parameters which have to be specified by the decision taker are the asset returns $V_a$, which are stochastic and need to be tree-approximated, as well as the deterministic budget $B$. The stochastic decision variables, which will be calculated via numerical optimization solver include the current (investment) budget $b_a$, purchases $p_a$, as well as sales $s_a$ of each asset $a$ out of the given investment universe $\mathcal{A}$ at each stage $t$. This model represents the basic building block and can be arbitrarily extended to the needs of the decision taker, e.g. by integrating dynamic risk measures, see e.g. \cite{EichhornRoemisch2005}. 

However, the crucial part of the whole stochastic programming workflow is to generate a multi-stage scenario tree, i.e. the $V_a$, which represents a careful approximation of the uncertainty of the asset (or asset category) returns, such that a sensible risk management can be based on it. 

This paper is organized as follows. Multi-stage scenario tree generation will be briefly sketched in Section \ref{ref:treegen}. A new evolutionary algorithm to create multi-stage scenario trees is presented in Section \ref{ref:evotreegen}. Section \ref{ref:numres} summarizes selected numerical results and the implementation, while Section \ref{sec:conclusion} concludes the paper.

\section{Multi-stage scenario tree generation}
\label{ref:treegen}

That scenario tree should represent the uncertain structure of the reality as close as possible, because the quality of the tree severely affects the quality of the solution of the multi-stage stochastic decision model, such that any approximation scheme should be done in consideration of some optimality criteria, i.e. before a stochastic optimization model is solved, a scenario optimization problem has to be solved independently of the optimization model. 

In the context of scenario optimization, optimality can be defined as the minimization of the distance between the original (continuous or highly discrete) stochastic process and the approximated scenario tree. Choosing an appropriate distance may be based on subjective taste, e.g. moment matching as proposed by \cite{HoylandWallace2001}, selected due to theoretical stability considerations (see \cite{RachevRomisch2002} and \cite{HeitschEtAl2006}), which leads to probability metric minimization problems as shown by \cite{Pflug2001} and \cite{DupacovaEtAl2003}, or it may be predetermined by chosen approximation method, e.g. by using different sampling schemes like QMC in \cite{PennanenKoivu2005} or RQMC in \cite{Koivu2005}, see also \cite{Pennanen2009}. It is important to remark that once the appropriate distance has been selected, an appropriate heuristic to approximate the chosen distance has to be applied, which affects the result significantly. 

Single-stage scenario generation, i.e. an optimal approximation of a multi-variate probability distribution without any tree structure can be done via various sampling as well as clustering techniques. The real algorithmic challenge of multi-stage scenario generation is maintaining a tree structure while still minimizing the overall distance. Only in rare cases, this problem can be solved without the application of heuristics. See \cite{Hochreiter2009} for a general overview of algorithmic aspects of multi-stage scenario generation, and \cite{HochreiterPflug2007} for details on financial multi-stage scenario generation.

\section{Evolutionary multi-stage scenario tree generation}
\label{ref:evotreegen}

The list of successful applications of evolutionary algorithms for solving financial problems is quickly growing, see especially \cite{BrabazonO2008}, \cite{BrabazonO2009}, \cite{DangBEO2009}, \cite{BrabazonOD2008}, and the references therein. This motivates for creating an evolutionary algorithm for the process of optimal multi-stage stochastic financial scenario generation. 

We assume that there is a finite set $\mathcal{S}$ of multi-stage, multi-variate scenario paths, which are sampled using the preferred scenario sampling engine selected by the decision taker. Stages will be denoted by $t = 1, \ldots, T$ where $t=1$ represents the (deterministic) root stage (root node), and $T$ denotes the terminal stage. Therefore, the input consists of a scenario path matrix of size $\vert \mathcal{S} \vert \times (T-1)$. Furthermore, the desired number of nodes of the tree in each stage is required, i.e. a vector $n$ of size $(T-1)$.

For the rest of the paper we will focus on the uni-variate case. However, the extension to the multi-variate case does not pose any structural difficulties besides that a dimension-weighting function for calculating the total distance on which the optimality of the scenario tree approximation is based on has to be defined.

A crucial part in designing a multi-stage scenario tree generator based on evolutionary techniques is finding a scalable genotype representation of a tree - both in terms of the numbers of stages as well as the number of input scenarios. The approach taken in this paper is using a real-valued vector in the range $[0,1]$ and mapping it to a scenario tree given the respective node format $n$. The length of the vector is equal to the number of input scenarios $s = \vert \mathcal{S} \vert$ plus the number of terminal nodes $n_T$. Thus, the presented algorithm is somewhat limited by the number of input scenarios. This means that input scenarios should not be a standard set of mindlessly sampled scenario paths, but rather a thoughtfully simulated view on the future uncertainty. This should not be seen as a drawback, as it draws attention to this often neglected part of the decision optimization process.

To map the real-valued vector to a scenario tree, which can be used for a subsequent stochastic optimization, two steps have to be fulfilled. First, the real-valued numbers are mapped to their respective node-set given the structure $n$ of the tree, and secondly, values have to be assigned to the nodes. There are different approaches to determine the center of the node-sets, which also affects the distance calculation, see below for more details.

It should be noted, that a random chromosome does not necessarily lead to a valid tree. This is the case if the number of mapped nodes is lower than the number of nodes necessary given by $n_t$ of the respective stage $t$. If an uniform random variable generator and a thoughtful node structure is used, which depends on the number of input scenarios, invalid trees should not appear frequently, and can be easily discarded if they do appear.

Consider the following example of the mapping procedure. For demonstration purposes, we only take one stage into account. We do have $10$ input scenarios (asset returns), each equipped with the same probability $p=0.1$, which might be the output of some sophisticated asset price sampling procedure, e.g.
$$(0.017,-0.023,-0.008,-0.022,-0.019,0.024,0.016,-0.006,0.032,-0.023).$$
We want to separate those values optimally into $2$ clusters, which then represent our output scenarios and take a random chromosome, which might look as follows:
$$(0.4387,0.3816,0.7655,0.7952,0.1869,0.4898,0.4456,0.6463,0.7094,0.7547)$$
If we map this vector to represent $2$ centers we obtain: $(1,1,2,2,1,1,1,2,2,2)$. Now we need to calculate a center value, e.g. the mean, and have to calculate the distance for each value of each cluster to its center, e.g. we obtain center means $(0.0032,-0.0055)$, which represent the resulting scenarios, each with a probability of $0.5$. The $l_1$ distance for each cluster is $(0.0975, 0.0750)$, so the objective function value is $0.1725$. Now flip-mutate chromosome $9$, i.e. $(1-0.7094)=0.2906$, such that input scenario $9$ (return = $0.032$) will now be part of cluster $1$ instead of cluster $2$. We obtain new scenarios $(0.0080, -0.0149)$ with probabilities $(0.6, 0.4)$. The objective function value is $0.1475$ (or $0.1646$ if you weight the distances with the corresponding output scenario probability), i.e. this mutation led to a better objective value.

It should be noted that this mapping is rather trivial in the single-stage case, but this simple approach leads to a powerful method for the tedious task of constructing multi-stage scenario trees for stochastic programming problems, because the nested probability structure of the stochastic process is implicitly generated.

\begin{figure}
\begin{center}
\scalebox{0.5}{
\includegraphics{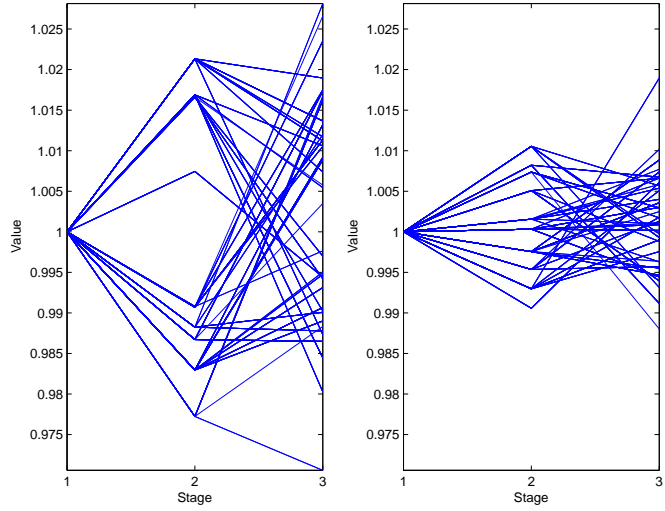}
}
\end{center}
\caption{Difference between Extreme (left) and Median (right) node-value mapping.}
\label{fig:medianextreme}
\end{figure}

For multi-stage trees, a crucial point is finding a representative value for the node-sets determined in the first step of the mapping. There exists a range of methods, which can be used for the determination of centers, i.e. mapping all values of a node-sets to one node. The distance of the approximation, which is used for the calculation of the objective function, will also be affected by this method. Some selected methods are summarized below:
\begin{itemize}
\item Median. A straight-forward solution is to use the median of the values.
\item Extreme. If the mean of the values is below the stage mean, the lowest value of the set will be selected, if it is above the stage mean, the highest value is selected. This can prove useful if one aims at capturing extremes, which already might have been flattened out by the scenario path simulation.
\item Mixture. Using the median approach might be smoothing the tail values too much, while the extreme approach neglects normal market phases. To overcome this, a mixture model can be defined, i.e. by splitting the range of stage values into three sections and using the minimum or maximum value of the node-set if the mean of it is in the lowest or highest section, or using the median if the node-set mean lies in the intermediate section.
\item Random. A randomly selected value will used. While this method generally leads to balanced results, decision takers might not favor this non-reproducible approach.
\end{itemize}
To visualize the differences of these approaches, see Fig. \ref{fig:medianextreme}. The same scenario generation procedure was used both for the left and the right part of the Figure, i.e. the same set of input scenarios, the same evolutionary algorithm parameters, and the same tree structure $n = [10,40]$. The only difference was choosing either the extreme node-to-value mapping (left) and the median mapping (right). It is clear that these two trees will lead to different decisions.
The set of input scenarios will be specified in detail in the next section, see the visualization of the scenario paths in Fig. \ref{fig:inputscen} below.

The evolutionary algorithm chosen is based on the commonly agreed standard as surveyed by \cite{BlumR03}. Thereby, the following evolutionary operators have been implemented and used:
\begin{itemize}
\item Elitist selection ($o_1$).
\item $N$-point crossover, with $N = 1$ ($o_2$) and $N = 2$ ($o_3$).
\item Intermediate crossover with a random intermediate probability, which is different for each chromosome ($o_4$).
\item Mutation/Flip: Invert an initially specified number of $m$ chromosomes by $1-c$, where $c$ is the current value ($o_5$).
\item Mutation/Random: An initially specified number of $m$ chromosomes will be randomly mutated ($o_6$).
\item Random addition: Randomly sampled chromosomes, also used for creating the initial population ($o_7$).
\end{itemize}
For each of crossover operator, one parent is taken from the $o_8$\% best of the previous population, and one entirely randomly. For each mutation operation one of the best $o_9$\% will be randomly used. These nine values $o_1, \ldots, o_9$ will be used for the description of numerical results and specify the percentage of the given population size, e.g. $(20,10,10,10,15,15,20,10,30)$ means that $20\%$ of each new population are created by applying elitist selection and random addition, while $10\%$ of each new population are created by crossovers ($1$-point crossover, $2$-point crossover, intermediate crossover) and $10\%$ by mutations (flip as well as random), where the crossovers are conducted with one parent randomly selected from the top $10\%$ and the other parent randomly selected from the whole previous population and the mutation is executed on one the top $30\%$ chromosomes from the previous population.

\section{Numerical results}
\label{ref:numres}

\begin{figure}
\begin{center}
\scalebox{0.5}{
\includegraphics{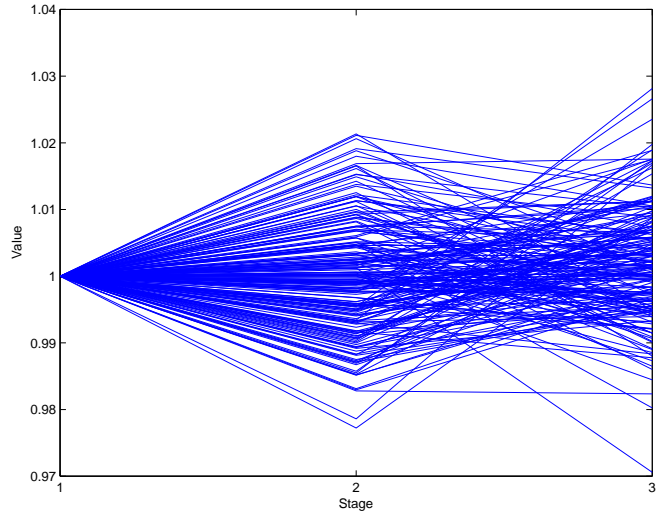}
}
\end{center}
\caption{The set of input scenarios used for numerical results $(s = 200)$.}
\label{fig:inputscen}
\end{figure}

The code was implemented using MatLab 2008b without using further toolboxes. Input scenarios have been estimated and simulated using a GARCH(1,1) time series model using historical data from the NASDAQ composite index. The input scenarios are shown in Fig. \ref{fig:inputscen}.

The results presented below have been calculated with the following parameters: The initial population consists of $1000$ randomly selected chromosomes. The population size during the evolutionary process has been set to $300$, and a maximum of $300$ iterations is calculated, using the above set of $200$ scenarios in two stages. The mutation parameter $m$ was set to $2$. The tree structure has been $[10,40]$ for all runs and each run takes around $4-5$ minutes to solve on an up-to-date desktop computer, which is excellent, when compared to other heuristic global optimization techniques, which often report a scenario generation time of many hours of computation.

\begin{figure}
\begin{center}
\begin{tabular}{cc}
\scalebox{0.3}{
\includegraphics{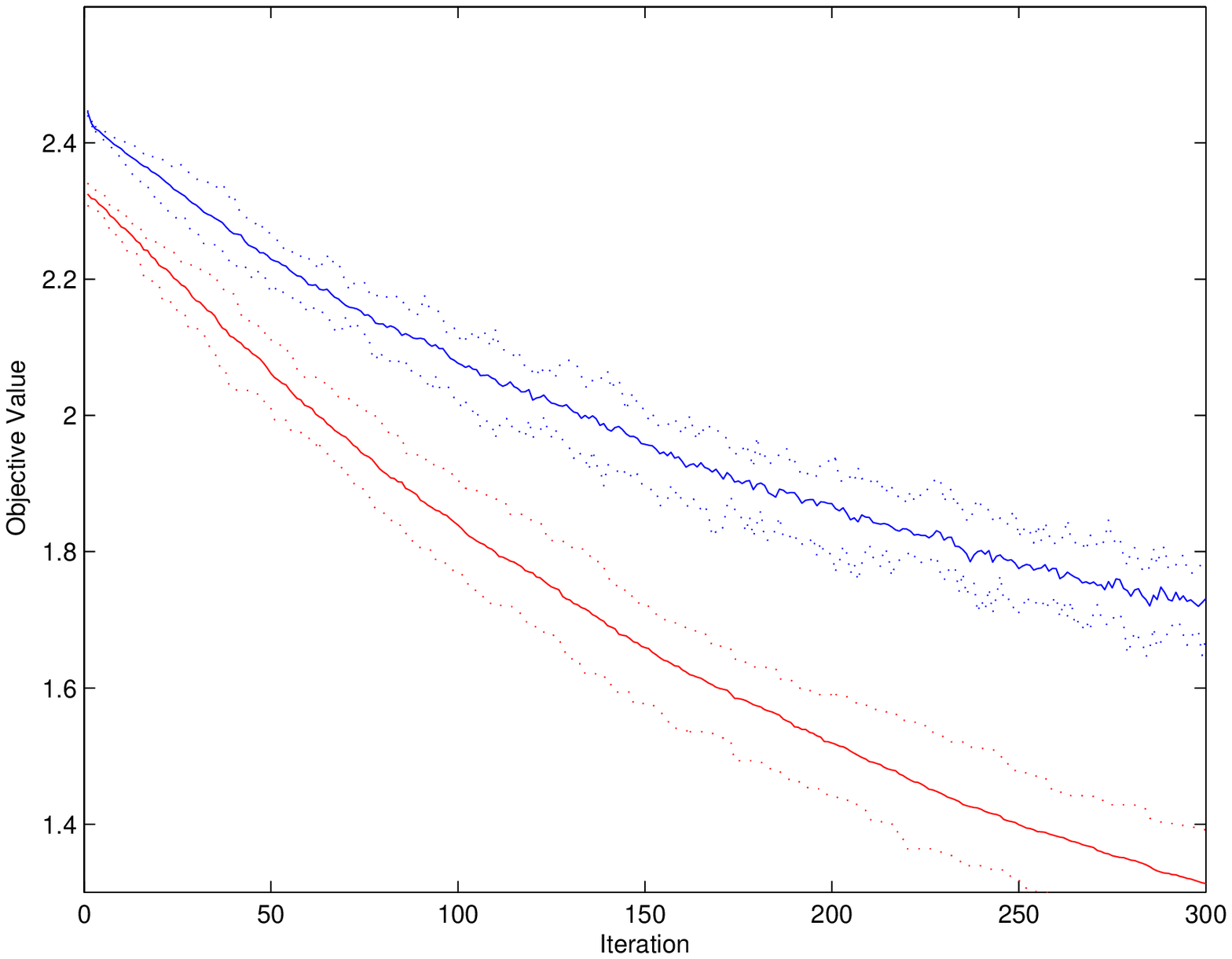}
}
&
\scalebox{0.3}{
\includegraphics{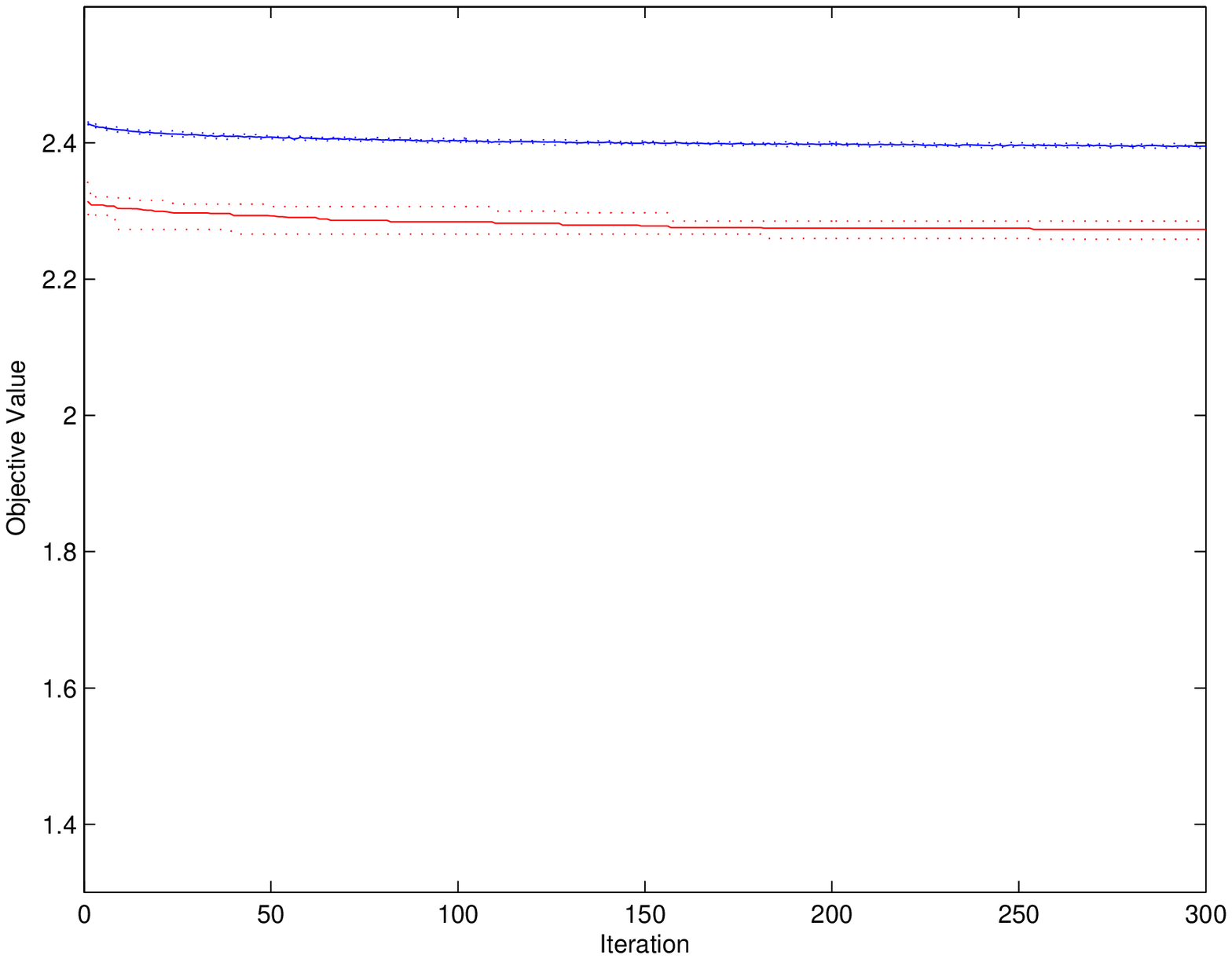}
}
\end{tabular}
\end{center}
\caption{Convergence of operator structure $(20,10,10,20,10,10,20,10,30)$ (left) and $(50,0,0,0,0,0,50,10,30)$ (right).}
\label{fig:convergence1}
\end{figure}

\begin{figure}
\begin{center}
\begin{tabular}{cc}
\scalebox{0.3}{
\includegraphics{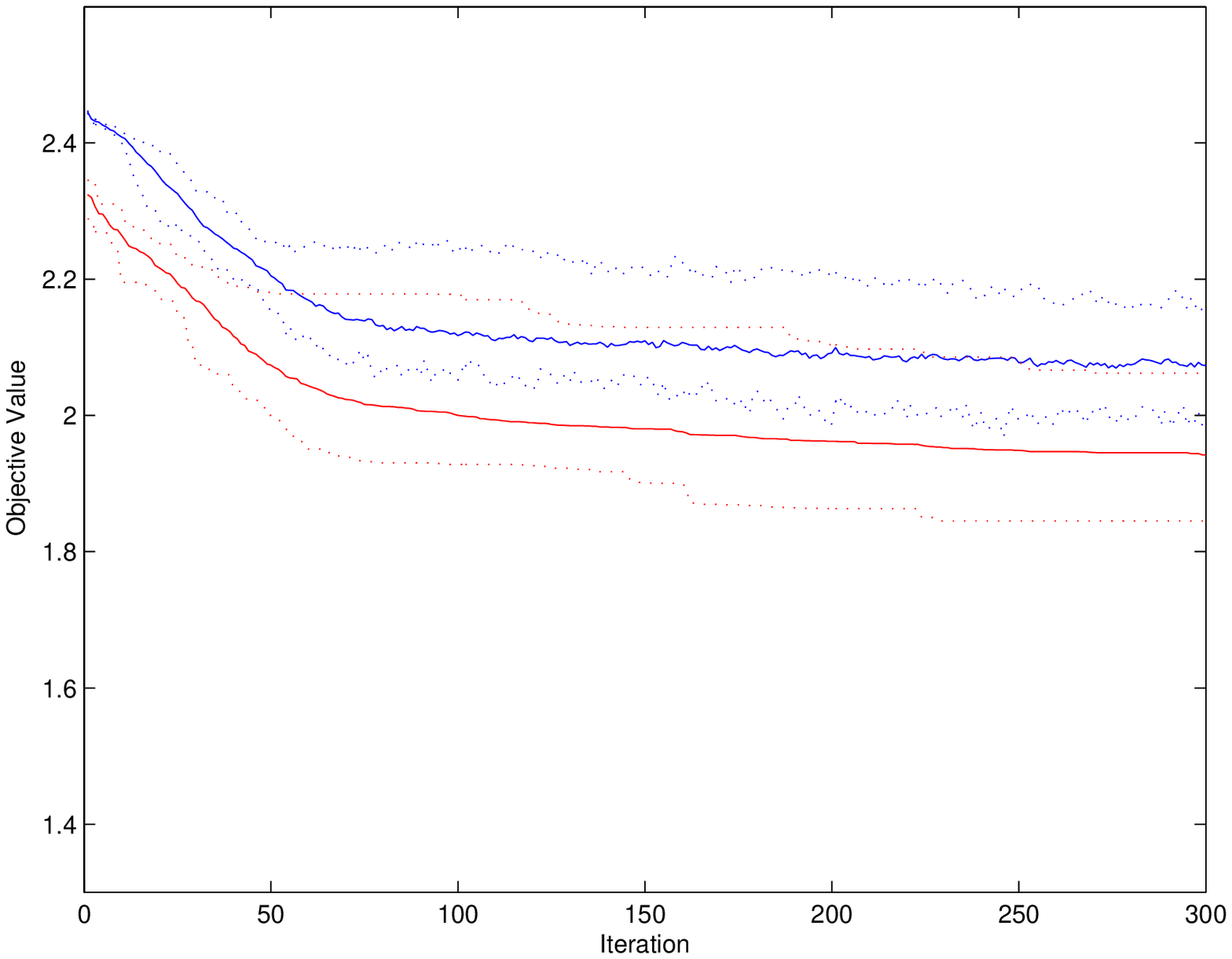}
}
&
\scalebox{0.3}{
\includegraphics{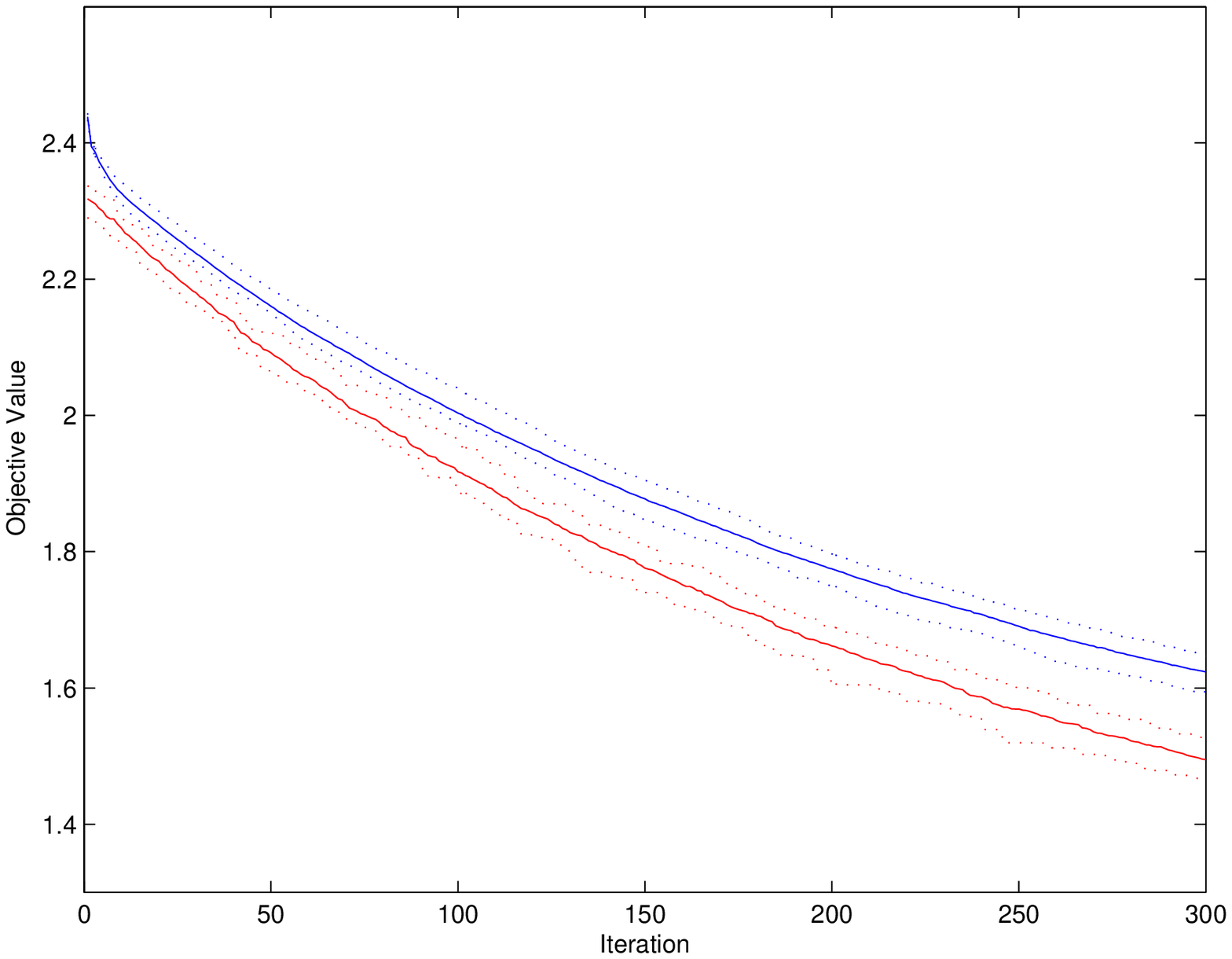}
}
\end{tabular}
\end{center}
\caption{Convergence of operator structure $(20,20,20,30,0,0,10,10,30)$ (left) and $(30,0,0,0,30,30,10,10,30)$ (right).}
\label{fig:convergence2}
\end{figure}

The first results show the convergence of different evolutionary operators. We compare four different operator structures:
\begin{itemize}
\item Using all operators $= (20,10,10,20,10,10,20,10,30)$, see Fig. \ref{fig:convergence1},
\item no crossover nor mutation $= (50,0,0,0,0,0,50,10,30)$, see Fig. \ref{fig:convergence1},
\item no mutation operators $= (20,20,20,30,0,0,10,10,30)$ see Fig. \ref{fig:convergence2}, and
\item no crossover operators $= (30,0,0,0,30,30,10,10,30)$ see Fig. \ref{fig:convergence2}.
\end{itemize}

The convergence graphs contain the minimum objective function value as well as the population mean. Each test has been repeated $10$ times, and the graphs show the mean of the two values, as well as the minimum and maximum per iteration. 

\begin{figure}
\begin{center}
\begin{tabular}{cc}
\scalebox{0.3}{
\includegraphics{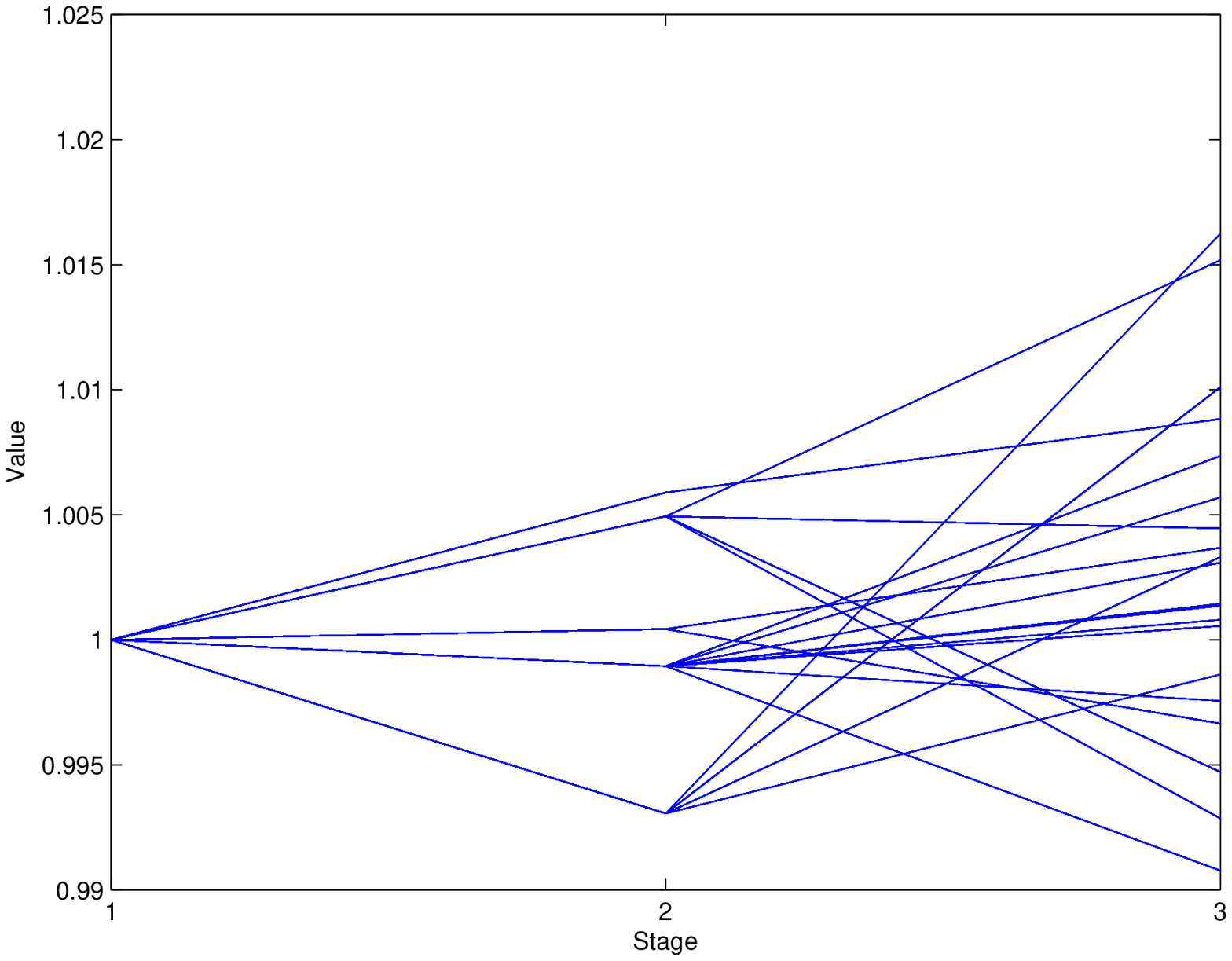}
}
&
\scalebox{0.3}{
\includegraphics{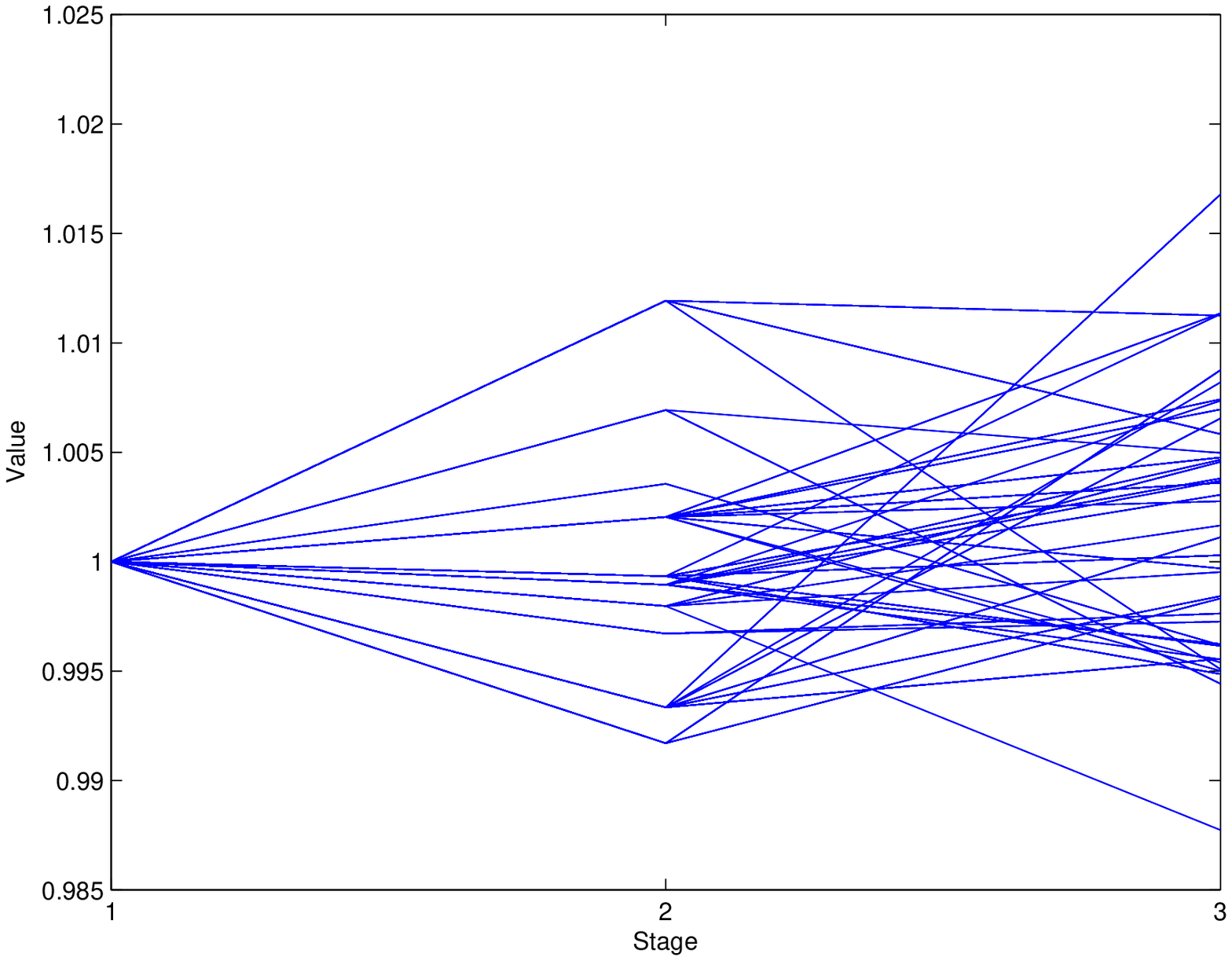}
}
\\
\scalebox{0.3}{
\includegraphics{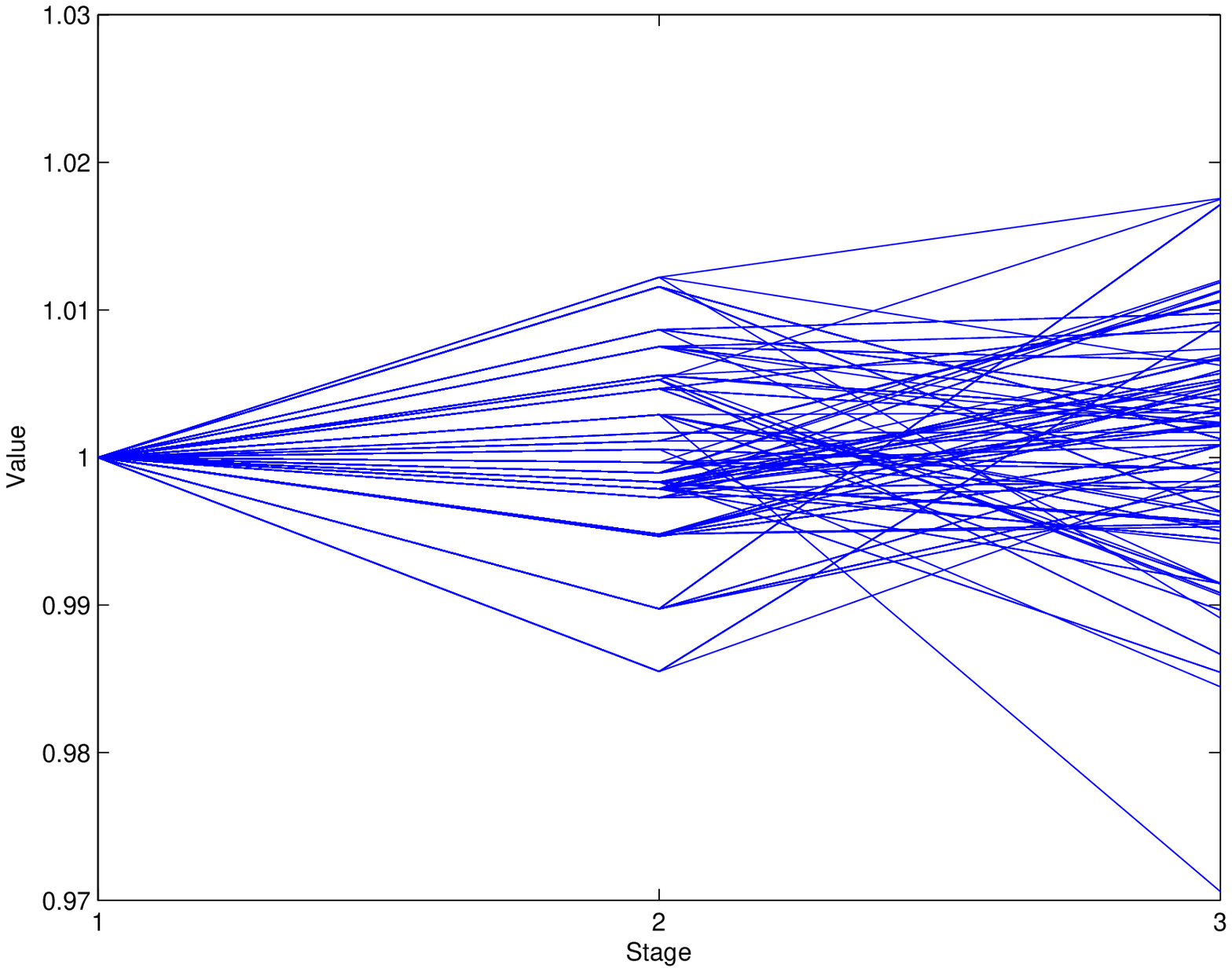}
}
&
\scalebox{0.3}{
\includegraphics{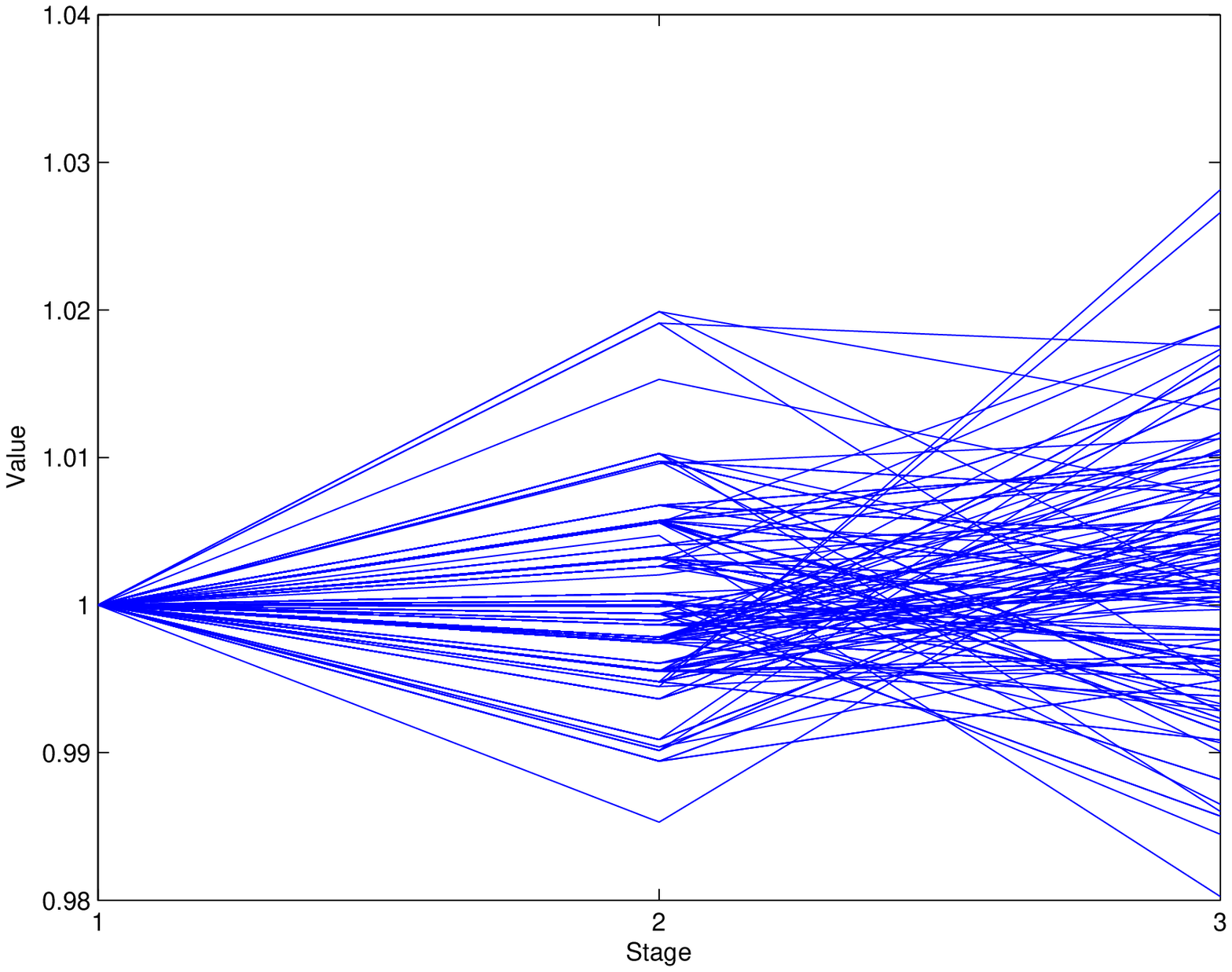}
}
\end{tabular}
\end{center}
\caption{Scenario trees with $n = [5,20], [10,40], [20,80],$ and $[40,120]$.}
\label{fig:treesize}
\end{figure}

In all calculations above, the same tree structure has been used, i.e. $n = [10,40]$. Of course, the method works for arbitrary scenario trees structures as shown in Fig. \ref{fig:treesize} for trees with a structure of $n = [5,20]$, $n = [10,40]$, $n = [20,80]$, and $n = [40,120]$ respectively. It should be noted that for any realistic application the evolutionary parameters have to be adapted to the specific instance of the scenario tree needed for the given stochastic optimization model.

\section{Conclusion}
\label{sec:conclusion}

In this paper an evolutionary multi-stage scenario tree generation method has been presented. It could be shown that multi-stage financial scenario generation can be successfully done by applying pure evolutionary optimization techniques. The results motivate for an extension of the implemented code for multi-variate scenario input paths and other features.

\bibliographystyle{splncs}
\bibliography{evofin2010}

\begin{thebibliography}{10}

\bibitem{RuszczynskiShapiro2003}
Ruszczy{\'n}ski, A., Shapiro, A., eds.:
\newblock Stochastic programming. Volume~10 of Handbooks in Operations Research
  and Management Science.
\newblock Elsevier Science B.V., Amsterdam (2003)

\bibitem{WallaceZiemba2005}
Wallace, S.W., Ziemba, W.T., eds.:
\newblock Applications of stochastic programming. Volume~5 of MPS/SIAM Series
  on Optimization.
\newblock Society for Industrial and Applied Mathematics (SIAM) (2005)

\bibitem{Markowitz1952}
Markowitz, H.M.:
\newblock Portfolio selection.
\newblock The Journal of Finance \textbf{7}(1) (1952)  77--91

\bibitem{Steinbach2001}
Steinbach, M.C.:
\newblock Markowitz revisited: mean-variance models in financial portfolio
  analysis.
\newblock SIAM Review \textbf{43}(1) (2001)  31--85

\bibitem{ArtznerEtAl1999}
Artzner, P., Delbaen, F., Eber, J.M., Heath, D.:
\newblock Coherent measures of risk.
\newblock Mathematical Finance \textbf{9}(3) (1999)  203--228

\bibitem{EichhornRoemisch2005}
Eichhorn, A., R{\"o}misch, W.:
\newblock Polyhedral risk measures in stochastic programming.
\newblock SIAM Journal on Optimization \textbf{16}(1) (2005)  69--95

\bibitem{HoylandWallace2001}
H{\o}yland, K., Wallace, S.W.:
\newblock Generating scenario trees for multistage decision problems.
\newblock Management Science \textbf{47}(2) (2001)  295--307

\bibitem{RachevRomisch2002}
Rachev, S.T., R{\"o}misch, W.:
\newblock Quantitative stability in stochastic programming: the method of
  probability metrics.
\newblock Mathematics of Operations Research \textbf{27}(4) (2002)  792--818

\bibitem{HeitschEtAl2006}
Heitsch, H., R{\"o}misch, W., Strugarek, C.:
\newblock Stability of multistage stochastic programs.
\newblock SIAM Journal on Optimization \textbf{17}(2) (2006)  511--525

\bibitem{Pflug2001}
Pflug, G.C.:
\newblock Scenario tree generation for multiperiod financial optimization by
  optimal discretization.
\newblock Mathematical Programming \textbf{89}(2, Ser. B) (2001)  251--271

\bibitem{DupacovaEtAl2003}
Dupa{\v{c}}ov{\'a}, J., Gr{\"o}we-Kuska, N., R{\"o}misch, W.:
\newblock Scenario reduction in stochastic programming. {A}n approach using
  probability metrics.
\newblock Mathematical Programming \textbf{95}(3, Ser. A) (2003)  493--511

\bibitem{PennanenKoivu2005}
Pennanen, T., Koivu, M.:
\newblock Epi-convergent discretizations of stochastic programs via integration
  quadratures.
\newblock Numerische Mathematik \textbf{100}(1) (2005)  141--163

\bibitem{Koivu2005}
Koivu, M.:
\newblock Variance reduction in sample approximations of stochastic programs.
\newblock Mathematical Programming \textbf{103}(3, Ser. A) (2005)  463--485

\bibitem{Pennanen2009}
Pennanen, T.:
\newblock Epi-convergent discretizations of multistage stochastic programs via
  integration quadratures.
\newblock Mathematical Programming \textbf{116}(1-2, Ser. B) (2009)  461--479

\bibitem{Hochreiter2009}
Hochreiter, R.:
\newblock Algorithmic aspects of scenario-based multi-stage decision process
  optimization.
\newblock In Rossi, F., Tsouki{\`a}s, A., eds.: Algorithmic Decision Theory
  2009. Volume 5783 of Lecture Notes in Computer Science., Springer (2009)
  365--376

\bibitem{HochreiterPflug2007}
Hochreiter, R., Pflug, G.C.:
\newblock Financial scenario generation for stochastic multi-stage decision
  processes as facility location problems.
\newblock Annals of Operations Research \textbf{152}(1) (2007)  257--272

\bibitem{BrabazonO2008}
Brabazon, A., O'Neill, M., eds.:
\newblock Natural Computing in Computational Finance. Volume 100 of Studies in
  Computational Intelligence.
\newblock Springer (2008)

\bibitem{BrabazonO2009}
Brabazon, A., O'Neill, M., eds.:
\newblock Natural Computing in Computational Finance, Volume 2. Volume 185 of
  Studies in Computational Intelligence.
\newblock Springer (2009)

\bibitem{DangBEO2009}
Dang, J., Brabazon, A., Edelman, D., O'Neill, M.:
\newblock An introduction to natural computing in finance.
\newblock In: EvoWorkshops 2009. Volume 5484 of Lecture Notes in Computer
  Science., Springer (2009)  182--192

\bibitem{BrabazonOD2008}
Brabazon, A., O'Neill, M., Dempsey, I.:
\newblock An introduction to evolutionary computation in finance.
\newblock IEEE Computational Intelligence Magazine \textbf{3}(4) (2008)  42--55

\bibitem{BlumR03}
Blum, C., Roli, A.:
\newblock Metaheuristics in combinatorial optimization: Overview and conceptual
  comparison.
\newblock ACM Computing Surveys \textbf{35}(3) (2003)  268--308

\end{thebibliography}

\end{document}